\documentclass[runningheads]{llncs}
\usepackage{combelow}
\usepackage{graphicx}

\usepackage{booktabs}
\usepackage{multirow}

\newcommand{\mycomment}[1]{}
\begin{document}
\title{RoBERTweet: A BERT Language Model for Romanian Tweets}

%

\author{Iulian-Marius T\u{a}iatu\textsuperscript{\rm 1,*} \and  Andrei-Marius Avram\textsuperscript{\rm 1,*} \and Dumitru-Clementin Cercel\textsuperscript{\rm 1} \and Florin Pop\textsuperscript{\rm 1,\rm2}}

\authorrunning{I.-M. T\u{a}iatu et al.}
%

\institute{
    \textsuperscript{\rm 1}Faculty of Automatic Control and Computers, University Politehnica of Bucharest \\
    \textsuperscript{\rm 2}National Institute for Research and Development in Informatics - ICI Bucharest, Romania \\
\email{\{iulian.taiatu,andrei\_marius.avram\}@stud.acs.upb.ro\\
\{dumitru.cercel, florin.pop\}@upb.ro}
}

\maketitle
\def\thefootnote{*}\footnotetext{Equal contribution}\def\thefootnote{\arabic{footnote}}
\begin{abstract}
Developing natural language processing (NLP) systems for social media analysis remains an important topic in artificial intelligence research. This article introduces RoBERTweet, the first Transformer architecture trained on Romanian tweets. Our RoBERTweet comes in two versions, following the base and large architectures of BERT. 
The corpus used for pre-training the models represents a novelty for the Romanian NLP community and consists of all tweets collected from 2008 to 2022. 
Experiments show that RoBERTweet models outperform the previous general-domain Romanian and multilingual language models on three NLP tasks with tweet inputs: emotion detection, sexist language identification, and named entity recognition. 
We make our models\footnote{\url{https://huggingface.co/Iulian277/ro-bert-tweet}} and the newly created corpus\footnote{\url{https://huggingface.co/datasets/Iulian277/romanian-tweets}} of Romanian tweets freely available.  

\end{abstract}

\section{Introduction}

Over the past several years, there has been a substantial surge in interest and enthusiasm related to natural language processing (NLP) techniques for various social media-related tasks. These techniques  may include but are not limited to training NLP models for purposes such as sentiment analysis, identifying offensive or sexist language, performing part-of-speech (POS) tagging, or conducting named entity recognition (NER). Nguyen et al. \cite{nguyen-etal-2020-bertweet} have presented a BERT \cite{devlin-etal-2019-bert} language model specifically developed for analyzing English tweets. Concretely, their model has been trained on a large corpus of 850 million English tweets and has yielded impressive results in several NLP tasks relevant to Twitter text analysis, such as POS tagging, NER, and text classification (i.e., irony detection and sentiment analysis).

Various other BERT models designed to analyze tweets have been developed, with each model focusing on a different language. For example, Guo et al. \cite{guo-etal-2021-bertweetfr} have designed and trained a BERT model to analyze French tweets. Similarly, Koto et al. \cite{koto-etal-2021-indobertweet} have presented their own BERT model optimized to analyze Indonesian tweets, while P\'erez et al. \cite{perez-etal-2022-robertuito} have created a BERT model that is capable of handling Spanish tweets. Additionally, Zhang et al. \cite{zhang2022twhin} have introduced a highly sophisticated multi-lingual BERT model trained on a massive dataset of over 7 billion tweets in more than 100 languages.

This paper presents two versions of language models for the Romanian language that were pre-trained on a novel corpus of Romanian tweets. In short, we summarized our main contributions as follows:
\begin{itemize}
    \itemsep0em
    \item We collect and release a novel corpus of Romanian tweets from 2008 to 2022, consisting of 65M unique tweets. It is a significant contribution to the field, as no such resource existed for the Romanian language.
   \item We conduct a comprehensive set of experiments based on this newly-created corpus, further demonstrating the usefulness of our dataset and pre-trained models.
    \item We develop the first language models pre-trained on the tweet domain for the Romanian language under the name RoBERTweet-base and RoBERTweet-large. These models can be utilized with the Transformer library \cite{wolf-etal-2020-transformers}, and we expect them to provide strong baselines for future research and applications related to Romanian tweet NLP tasks.
    \item We obtain state-of-the-art (SOTA) results on three existing Romanian tweet tasks: emotion detection (ED), named entity recognition, and sexist language identification (SLI).
\end{itemize}

\section{Related Work}

The English BERT language model was pre-trained on two large datasets, BookCorpus and English Wikipedia, using two prediction tasks, masked language model (MLM) and next sentence prediction (NSP). After pre-training, BERT was fine-tuned on several target datasets where it learned to make inferences on specific tasks such as NER, question answering, or natural language inference, obtaining state-of-the-art results on the General Language Understanding Evaluation (GLUE) benchmark \cite{wang2018glue}. Since BERT was introduced, this mechanism of first pre-training a language model on a large corpus and then fine-tuning the resulting model on specific tasks has become ubiquitous in NLP. Thus,  many researchers have flocked to this area, developing larger and more complex language models.

One important iteration in this direction is the Robustely Optimized BERT (RoBERTa) model \cite{liu2020roberta}, which argues that the BERT model is sub-optimized. Therefore, RoBERTa pre-trained the architecture on a larger corpus with a larger batch size, applied the dynamic MLM, and removed the NSP task, achieving state-of-the-art performance on the GLUE benchmark. 
A Lite version of BERT (ALBERT) \cite{Lan2020ALBERT:} is another important architecture introduced in the literature that addressed the issue associated with many parameters of BERT and solved it by proposing a cross-layer parameter sharing and an embedding parameter factorization technique. To achieve better results, the Efficiently Learning an Encoder that Classifies Token Replacement Accurately (ELECTRA) \cite{clark2020electra} applied a replace token detection (RTD) task in pre-training instead of employing MLM.  

The first Romanian BERT models (i.e., BERT-base-cased-ro and BERT-base-uncased-ro) were developed by Dumitrescu et al. \cite{dumitrescu-etal-2020-birth}. The models were trained on a general-domain corpus collected from diverse sources such as OPUS \cite{tiedemann-2012-parallel}, Wikipedia dumps\footnote{\url{https://github.com/dumitrescustefan/wiki-ro}}, and OSCAR \cite{ortizsuarez:hal-02148693}, totalling over 15.2 GB of text data. Since then, various BERT-like models have been introduced in the research literature, including other general variants of the Romanian BERT (i.e., RoBERT-small, RoBERT-base, and RoBERT-large) \cite{masala2020robert}, the distilled versions of Romanian BERTs (i.e., Distil-BERT-base-ro, Distil-RoBERT-base, and DistilMulti-BERT-base-ro) \cite{avram-etal-2022-distilling}, the "judiciar"\footnote{"Judiciar" is the Romanian equivalent to the English "criminal record".} BERT (i.e., jurBERT-base and jurBERT-large) \cite{masala2021jurbert} trained on legal data, and A Lite Romanian BERT (ALR-BERT) \cite{nicolae2022lite}, a monolingual language model that follows the ALBERT architecture.

\section{RoBERTweet}

\subsection{Dataset}

\begin{figure}[h]
    \centering
    \includegraphics[width=0.95\textwidth]{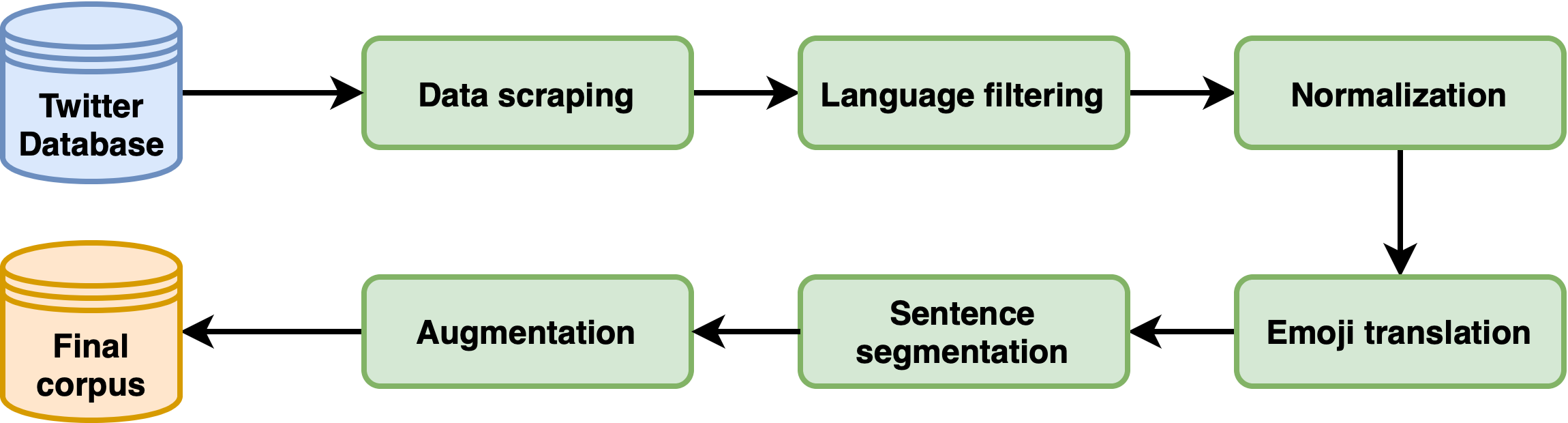}
    \caption{RoBERTweet training corpus pre-processing pipeline.}
    \label{fig:dataset}
\end{figure}

Figure \ref{fig:dataset} depicts the dataset generation pipeline. We crawl Romanian tweets over an extensive period from 2008 to 2022 using the \textit{snscrape} package\footnote{\url{https://github.com/JustAnotherArchivist/snscrape}}. When sending the query, we set the \textit{RO} flag to ensure the corpus contains only Romanian text. We also employ the language identification component of \textit{fastText} \cite{joulin-etal-2017-bag} and the \textit{langdetect}  library\footnote{\url{https://pypi.org/project/langdetect}} to further filter out non-Romanian tweets. These are crucial steps because of pre-processing is not handled accordingly, the resulting corpus would contain text in other languages that introduce noise and uncertainty to the model predictions while also potentially increasing the number of \texttt{[UNK]} tokens.

The following pre-processing steps are inspired by the existing approach of Nguyen et al. \cite{nguyen-etal-2020-bertweet}. 
For each tweet, we employ the normalization technique to convert user mentions, URL links, and hashtags into special tokens, namely \texttt{USER}, \texttt{HTTPURL}, and \texttt{HASHTAG}. We further exclude tweets shorter than five or longer than 256 words and filter out the tweets that contain more than three mentions, three hashtags, three URLs, or three emojis, as they are considered irrelevant or spam and may add noise to the training procedure. In addition, to increase the language understanding and make the model stronger for future downstream tasks (e.g., emotion detection), we translate each emoji with its corresponding text. This is done using the \textit{emoji} package\footnote{\url{https://pypi.org/project/emoji}}.

We also use the \textit{sentencizer} component from \textit{spaCy}\footnote{\url{https://spacy.io/api/sentencizer}} to split the tweets into sentences. This observation is necessary for the NSP task, which is part of the pre-training objective of the BERT model. After each sentence of a tweet, we introduce a new line, and after the last sentence of a tweet, we introduce two blank lines, thus delimiting two tweets.

After performing all these data cleaning and normalization steps, we obtained a 5GB corpus containing 65M tweets, which we open-sourced and made publicly available. Finally, we convert the uncompressed text files to \textit{tfrecords} using the script that creates the pre-training data\footnote{\url{https://github.com/google-research/bert}}. We augment the dataset using a dupe factor of 10, meaning that each tweet is randomly masked ten times using different seeds. Thus, we generate a tenfold increase in the size of our pre-training corpus, effectively performing a form of data augmentation.

\subsection{Methodology}

In this study, the developed RoBERTweet models use a cased tokenizer and have the same architecture as BERT-base and BERT-large. The pre-training procedure is also the same, which involves using MLM and NSP objectives. The MLM objective selects 15\% of the tokens for possible replacement, with 80\% of those selected being replaced by the special [MASK] token, 10\% remaining unchanged, and 10\% replaced by random tokens from the vocabulary. The NSP objective involves pairing 50\% of the input sentences with the real subsequent sentence as the second sentence. In contrast, in the other 50\% of cases, a random sentence from the corpus is selected as the second sentence.

\begin{table}
    \centering

    \caption{ Comparison of our RoBERTweet models and the other Romanian BERT models available in the literature regarding their training data size, number of layers, number of hidden units on each layer, number of heads, vocabulary dimension, and number of parameters.}
    
    \begin{tabular}{|l|r|r|r|r|r|r|}
         \toprule
         \textbf{Model} & \textbf{Train Size} & \textbf{Layers} & \textbf{Hidden} & \textbf{Heads} & \textbf{Vocab} & \textbf{Params} \\
         \midrule
         BERT-base-cased-ro \cite{dumitrescu-etal-2020-birth} & 15.2 GB & 12 & 768 & 12 & 50K & 124M \\
         BERT-base-uncased-ro \cite{dumitrescu-etal-2020-birth} & 15.2 GB & 12 & 768 & 12 & 50K & 124M \\
         RoBERT-small \cite{masala2020robert} & 12.6 GB & 12 & 256 & 8 & 38K & 19M \\
         RoBERT-base \cite{masala2020robert} & 12.6 GB & 12 & 768 & 12 & 38K & 114M \\
         RoBERT-large \cite{masala2020robert} & 12.6 GB & 24 & 1024 & 16 & 38K & 341M \\
         Distil-BERT-base-ro \cite{avram-etal-2022-distilling} & 25.3 GB & 6 & 768 & 12 & 50K & 81M \\
         Distil-RoBERT-base \cite{avram-etal-2022-distilling} & 25.3 GB & 6 & 768 & 12 & 38K & 72M \\
         DistilMulti-BERT-base-ro \cite{avram-etal-2022-distilling} & 25.3 GB & 6 & 768 & 12 & 50K & 81M \\
         jurBERT-base \cite{masala2021jurbert} & 160 GB & 12 & 768 & 12 & 33K & 111M \\
         jurBERT-large \cite{masala2021jurbert} & 160 GB & 24 & 1024 & 16 & 33K & 81M \\
         ALR-BERT \cite{nicolae2022lite} & 15.2 GB & 12 & 768 & 12 & 50K & 81M \\
         \midrule
        RoBERTweet-base (ours) & 5 GB & 6 & 768 & 12 & 51K & 110M \\
        RoBERTweet-large (ours) & 5 GB & 6 & 768 & 12 & 51K & 345M \\
         \bottomrule
    \end{tabular}
    \label{tab:model_size}
\end{table}

The weights are randomly initialized, and training is initiated from scratch. The base version (i.e., RoBERTweet-base) uses 12 layers, 768 hidden dimensions, and 12 attention heads, resulting in 110M parameters. In comparison, the large version (i.e., RoBERTweet-large) has 24 layers, 1024 hidden dimensions, and 16 attention heads, resulting in 345M parameters.
Inspired by Zampieri et al. \cite{zampieri2023language}, we depict in Table  \ref{tab:model_size} more details regarding the architecture of our RoBERTweet models compared to the other available Romanian BERT models.

In language modeling, the vocabulary plays a critical role in model performance. Generally speaking, the better the tokenization of sentences (i.e., the fewer pieces each word is broken into), the better the model is expected to perform. In this study, the vocabulary used for the models is an extension of the BERT-base-ro vocabulary, created by adding the special tweet tokens resulting from pre-processing (i.e., \texttt{USER}, \texttt{HTTPURL}, and \texttt{HASHTAG}) and the most frequent 25\% of emojis found in the tweets. The resulting vocabulary contained 51K tokens, an increment of 1K tokens compared to the vocabulary used by BERT-base-ro.

Each RoBERTweet variant was trained using a v2-8 Tensor Processing Unit (TPU), requiring approximately three weeks each.
The RoBERTweet models were trained using a batch size of 2048 and a total of 3M steps. However, the pre-training process was stopped after 1.5M steps since the model ceased to make any progress. We employed the Adam optimizer \cite{Kingma2014AdamAM}, together with a linear scheduler. The first 30K steps (i.e., 1\% of the total steps) are used for warming up the learning rate.

\section{Experiments and Results}

Since our RoBERTweet models are the first pre-trained language models for Romanian tweets, we compare them with the Romanian general-domain BERT language models\footnote{We were not able to load the ALR-BERT model from the HuggingFace repository (available at \url{https://huggingface.co/dragosnicolae555/ALR\_BERT}), so we did not include its results in our evaluation.}: BERT-base-cased-ro, BERT-base-uncased-ro, RoBERT-small, RoBERT-base, RoBERT-large, DistilMulti-BERT-base-ro, Distil-BERT-base-ro, and Distil-RoBERT-base. Subsequently, we fine-tune and analyze the results of our RoBERTweet models and the other general-domain BERT language models on three Twitter-related NLP tasks: emotion detection, sexist language identification, and named entity recognition.

\subsection{Emotion Detection}

The ED task is a supervised classification task that aims to predict the emotion of a given text. To perform this task, we use the second version of the Romanian emotion dataset (REDv2) \cite{redv2}, a helpful resource designed for detecting emotions in Romanian tweets. It is an extension of the REDv1 dataset \cite{RED} and consists of 5,449 multi-label annotated tweets with seven types of emotions: anger, fear, happiness, sadness, surprise, trust, and neutral. We present the results of the evaluated models treating this task as either classification or regression.

Following the methodology proposed by Devlin et al. \cite{devlin-etal-2019-bert}, we append a linear prediction layer on top of the mean output of the pre-trained language model and include a dropout rate of 10\% for regularization. For the fine-tuning process, we utilize the AdamW optimizer \cite{loshchilov2018decoupled} with a fixed learning rate of 2e-5 and a batch size of 16. To prevent overfitting, we implement early stopping, which stops the training procedure when no performance improvement happens on the validation set in 3 consecutive epochs. We compute accuracy (Acc), the Hamming loss (Ham), the F1-score (F1), and the mean squared error (MSE) score on the test set for each model. 

The classification and regression results for ED are depicted in Table \ref{tab:sa}. Except for MSE, the highest scores were obtained by RoBERTweet-large, achieving a Hamming loss of 0.085, an accuracy of 58.6\%, and an F1-score of 69.6\% when the ED task was treated as classification, and a Hamming loss of 0.088, an accuracy of 57.6\%, and an F1-score of 70.0\% when the ED task was treated as regression. It outperforms its general domain counter-part, RoBERT-large, on all these metrics, improving the classification Hamming loss by 0.003, accuracy by 0.8\%, and F1-score by 0.5\%, as well as the regression Hamming loss by 0.007, accuracy by 2.4\%, and F1-score by 1.3\%.  

\begin{table*}
    \centering
    \caption{Emotion detection scores in the classification and regression setting. The baseline models in the paper that introduced the dataset are depicted in italics.}
    \label{tab:sa}
    \begin{tabular}{|l|c|c|c|c|c|c|c|c|}
        \toprule
        \multirow{ 2}{*}{\textbf{Model}} & \multicolumn{4}{c|}{\textbf{\underline{Classification}}} & \multicolumn{4}{c|}{\textbf{\underline{Regression}}} \\
         & \textbf{Ham} & \textbf{Acc} & \textbf{F1} & \textbf{MSE} & \textbf{Ham} & \textbf{Acc} & \textbf{F1} & \textbf{MSE} \\
         \midrule
         \textit{BERT-base-cased-ro} \cite{redv2} & 0.104 & 0.541 & 0.668 & 26.74 & 0.970 & 0.542 & 0.670 & 10.06 \\
         \textit{XLM-RoBERTa} \cite{redv2} & 0.121 & 0.504 & 0.619 & 18.40 & 0.104 & 0.522 & 0.649 & 9.56 \\
         BERT-base-cased-ro \cite{dumitrescu-etal-2020-birth} & 0.105 & 0.549 & 0.675 & 20.87 & 0.098 & 0.541 & 0.664 & 10.33 \\
         BERT-base-uncased-ro \cite{dumitrescu-etal-2020-birth} & 0.097 & 0.547 & 0.669 & 9.83 & 0.097 & 0.546 & 0.679 & 9.92 \\
         RoBERT-small \cite{masala2020robert} & 0.106 & 0.524 & 0.642 & 9.93 & 0.102 & 0.529 & 0.645 & 7.62 \\
         RoBERT-base \cite{masala2020robert} & 0.106 & 0.550 & 0.666 & 26.79 & 0.094 & 0.566 & 0.684 & 9.67 \\
         RoBERT-large \cite{masala2020robert} & 0.098 & 0.578 & 0.691 & 18.49 & 0.095 & 0.552 & 0.687 & 9.50 \\
         Distil-BERT-base-ro \cite{avram-etal-2022-distilling} & 0.113 & 0.502 & 0.623 & 14.47 & 0.106 & 0.488 & 0.627 & 8.81 \\
         Distil-RoBERT-base \cite{avram-etal-2022-distilling} & 0.101 & 0.550 & 0.659 & \textbf{6.85} & 0.096 & 0.570 & 0.653 & \textbf{6.41} \\
         DistilMulti-BERT-base-ro \cite{avram-etal-2022-distilling} & 0.107 & 0.552 & 0.661 & 15.55 & 0.105 & 0.550 & 0.648 & 7.82 \\
   
         \midrule 
         RoBERTweet-base (ours) & 0.102 & 0.556 & 0.677 & 22.60 & 0.096 & 0.557 & 0.673 & 9.92 \\
         RoBERTweet-large (ours) & \textbf{0.095} & \textbf{0.586} & \textbf{0.696} & 24.43 & \textbf{0.088} & \textbf{0.576} & \textbf{0.700} & 9.67 \\
         \bottomrule
    \end{tabular}
\end{table*}

The lowest MSE scores were obtained by Distil-RoBERT-base, with a 6.85 MSE for classification ED and 6.41 MSE for regression ED. To the best of our knowledge, we do not have a clear justification of why this model specifically performed so well on this metric, and a detailed analysis would be outside the scope of this paper. Our only observations are: (1) the classification MSE scores have a high variance ranging from 6.85 to 26.79 with no consistent pattern in the results, and (2) the MSE regression scores are lower than the classification MSE scores. These two observations are expected since, compared to regression, we do not try to minimize the MSE loss in classification directly.

\subsection{Sexist Language Identification}

The CoRoSeOf corpus \cite{hoefels-etal-2022-coroseof} is a 	manually annotated resource that aims to identify the sexist and offensive language in Romanian tweets. The dataset contains approximately 40K tweets and five labels: sexist direct, sexist descriptive, sexist reporting, non-sexist offensive, and non-sexist non-offensive. Using these labels, two evaluation tasks were derived from this dataset: binary classification, where a model has to predict whether a tweet is a sexist or non-sexist, and three-way classification, where a model has to predict the kind of sexism in a tweet identified as such (i.e., direct, descriptive, and reporting).

For this experiment, we adopt the same fine-tuning strategy as in the previous task, but our target is to maximize the F1-score in this task. We compute the precision, recall, and F1-scores of our RoBERTweet variants and compare them with the scores achieved by the other general-domain Romanian BERT models available in the literature. In addition, we outline the results of the Support Vector Machine (SVM) \cite{cortes1995support} baseline model introduced in the dataset paper \cite{hoefels-etal-2022-coroseof}.

\begin{table*}
    \centering
    \caption{Sexist language identification results for both the binary and three-way classification tasks. The baseline models in the paper that introduced the dataset are depicted in italics.}
    \label{tab:sli}
    \begin{tabular}{|l|c|c|c|c|c|c|}
        \toprule
        \multirow{ 2}{*}{\textbf{Model}} & \multicolumn{3}{c|}{\textbf{\underline{Binary}}} & \multicolumn{3}{c|}{\textbf{\underline{Three-way}}} \\
         & \textbf{P} & \textbf{R} & \textbf{F1} & \textbf{P} & \textbf{R} & \textbf{F1} \\
         \midrule
         \textit{SVM} \cite{hoefels-etal-2022-coroseof} & 0.830 & 0.832 & 0.831 & 0.693 & 0.700 & 0.716 \\
         BERT-base-cased-ro \cite{dumitrescu-etal-2020-birth} & 0.844 & 0.812 & 0.836 & 0.821 & 0.753 & 0.778 \\
         BERT-base-uncased-ro \cite{dumitrescu-etal-2020-birth} & 0.846 & 0.814 & 0.829 & 0.878 & 0.732 & 0.771 \\
         RoBERT-small \cite{masala2020robert} & 0.855 & 0.811 & 0.831 & 0.840 & 0.662 & 0.677 \\
         RoBERT-base \cite{masala2020robert} & 0.849 & 0.784 & 0.812 & 0.812 & 0.710 & 0.736 \\
         RoBERT-large \cite{masala2020robert} & 0.853 & 0.833 & 0.842 & \textbf{0.894} & 0.740 & 0.777 \\
         Distil-BERT-base-ro \cite{avram-etal-2022-distilling} & 0.828 & 0.795 & 0.810 & 0.730 & 0.654 & 0.673 \\
         Distil-RoBERT-base \cite{avram-etal-2022-distilling} & 0.855 & 0.753 & 0.793 & 0.795 & \textbf{0.763} & 0.777 \\
         DistilMulti-BERT-base-ro \cite{avram-etal-2022-distilling} & 0.860 & 0.747 & 0.790 & 0.708 & 0.610 & 0.655 \\
         
         \midrule 
         RoBERTweet-base (ours) & \textbf{0.866} & 0.821 & 0.841 & 0.798 & 0.724 & 0.749 \\
         RoBERTweet-large (ours) & 0.837 & \textbf{0.878} & \textbf{0.855} & 0.823 & 0.754 & \textbf{0.780} \\
         \bottomrule
    \end{tabular}
\end{table*}

Table \ref{tab:sli} depicts our binary and three-way classification task results. The highest F1-scores for both binary and three-way SLI were obtained by RoBERTweet-large with 85.5\% and 78.0\%, respectively. It outperformed the second best model, RoBERT-large\footnote{Distil-RoBERT-base achieves the same F1-score on three-way SLI as RoBERTweet-large.}, by 1.2\% and 0.3\% on the binary and three-way SLI, respectively. RoBERTweet-base also performed well on this task, achieving a higher F1-score than all the other base and smaller Romanian models on binary SLI. However, the same does not hold on the three-way SLI classification, and it fell behind BERT-base-cased-ro, BERT-base-uncased-ro, and RoBERTweet-base on this subtask.

\subsection{Named Entity Recognition}

P\u{a}i\cb{s} et al. \cite{pais-etal-2022-romanian} have constructed a NER dataset from microblogging texts sourced from social media platforms (i.e., Twitter, Reddit, and Gab). The dataset contains high-quality annotations for nine entity types as follows: persons (PER), locations (LOC), organizations (ORG), time expressions (TM), legal references (LEG), disorders (DIS), chemicals (CHM), medical devices (MD), and anatomical parts (ANT). A total of 7,800 messages have been manually annotated in the dataset, which in turn contains 11K annotations. We use this dataset to train our models to perform NER on Romanian microblogging. 

Fine-tuning follows the supervised token classification method proposed by Devlin et al. in \cite{devlin-etal-2019-bert}. Specifically, we employ a fully-connected layer on top of the embeddings produced by BERT that correspond to the first subword token of each word. The models are trained using the AdamW optimizer with a fixed learning rate of 2e-5 and a batch size of 16, to which we add the weight decay. We compute performance scores for each entity class and the overall F1-score. The baselines used for comparison are those reported in the literature \cite{pais-etal-2022-romanian} and consist of three systems: Neuroner CoRoLa, Neuroner MB, and Neuroner CoRoLa + MB.

\setlength{\tabcolsep}{5.5pt}
\begin{table*}
    \centering
    \caption{NER performance for each entity class and the overall F1-score. The baseline models in the paper that introduced the dataset are depicted in italics.}
    \label{tab:ner}
    \resizebox{\textwidth}{!}{
    \begin{tabular}{|l|c|c|c|c|c|c|c|c|c|c|}
        \toprule
        \textbf{Model} & \textbf{ANT} & \textbf{CHM} & \textbf{DIS} & \textbf{LEG} & \textbf{LOC} & \textbf{MD} &  \textbf{ORG} & \textbf{PER} &  \textbf{TM} & \textbf{Total} \\
         \midrule
         \textit{Neuroner CoRoLa} \cite{pais-etal-2022-romanian} & 42.96 & 60.47 & 75.47 & 45.71 & 77.69 & 72.73 & 66.21 & 84.14 & 63.96 & 72.03 \\
         \textit{Neuroner MB} \cite{pais-etal-2022-romanian} & 22.54 & 58.82 & 71.43 & 47.37 & 81.27 & 61.54 & 65.95 & 80.95 & 63.64 & 71.26 \\
         \textit{Neuroner CoRoLa + MB} \cite{pais-etal-2022-romanian} & 21.43 & \textbf{82.87} & 73.47 & 36.36 & 81.21 & \textbf{66.67} & 62.00 & 83.51 & 61.50 & 70.75 \\
         BERT-base-cased-ro \cite{dumitrescu-etal-2020-birth} & 50.01 & 58.42 & 74.61 & 47.61 & 83.43 & 66.66 & \textbf{71.19} & 84.17 & 64.02 & 74.14 \\
         BERT-base-uncased-ro \cite{dumitrescu-etal-2020-birth} & \textbf{61.11} & 60.21 & 79.59 & 43.24 & 82.22 & 46.15 & 69.19 & 85.57 & 61.89 & 74.16 \\
         RoBERT-small \cite{masala2020robert} & 38.85 & 58.82 & 74.14 & 32.65 & 79.38 & 18.18 & 67.09 & 79.04 & 59.91 & 69.54 \\
         RoBERT-base \cite{masala2020robert} & 50.21 & 66.39 & 76.35 & \textbf{54.00} & 82.55 & 41.28 & 69.44 & \textbf{85.70} & 63.40 & 74.14 \\
         RoBERT-large \cite{masala2020robert} & 55.20 & 63.01 & \textbf{80.73} & 47.38 & 82.42 & 60.03 & 69.91 & 85.04 & \textbf{66.62} & 74.22 \\
         Distil-BERT-base-ro \cite{avram-etal-2022-distilling} & 42.85 & 60.21 & 70.46 & 47.61 & 77.02 & 30.76 & 63.17 & 75.28 & 62.97 & 68.51 \\
         Distil-RoBERT-base \cite{avram-etal-2022-distilling} & 36.11 & 53.33 & 68.65 & 40.90 & 75.00 & 30.76 & 60.37 & 75.52 & 60.30 & 66.31 \\
         DistilMulti-BERT-base-ro \cite{avram-etal-2022-distilling} & 41.77 & 33.45 & 46.61 & 51.22 & 72.22 & 60.03 & 44.77 & 72.47 & 56.91 & 65.69 \\
         \midrule 
         RoBERTweet-base (ours) & 45.33 & 69.66 & 77.94 & 50.00 & \textbf{83.84} & 57.14 & 70.29 & 83.31 & 64.90 & \textbf{74.43} \\
         RoBERTweet-large (ours) & 44.73 & 65.90 & 76.09 & 51.28 & 82.80 & 61.53 & 71.08 & 85.43 & 62.97 & 74.26 \\
         \bottomrule
    \end{tabular}
    }
\end{table*}

Table \ref{tab:ner} depicts our results of the evaluated models on NER. The highest F1-score was obtained by the RoBERTweet-base model with 74.43\%, RoBERTweet-large occupying the second place with an F1-score of 74.26\%. An interesting result of this evaluation is that except for the LOC entity, RoBERTweet-base and RoBERTweet-large do not obtain the best F1-score on individual entities. However, the two models get the first two best average F1-scores. This may indicate that general language models pay too much attention to certain entities due to their lack of domain-specific knowledge. In contrast, the performance of domain-specific language models is more evenly distributed between entities. However, further analysis is necessary to confirm this observation for other domains, models, and languages, which would be outside the scope of this paper.

\section{Conclusions}

In this work, we presented the first pre-trained language models for Romanian tweets on a large scale: RoBERTweet-base and RoBERTweet-large, together with the novel corpus on which they were trained. This corpus comprises 65M Romanian tweets which have been carefully filtered and normalized. Our experimental results demonstrated the usefulness of both RoBERTweet variants by outperforming the baselines, while also showing superior performance compared to the previous SOTA Romanian BERT models on three downstream Twitter NLP tasks: emotion detection, sexual language identification, and named entity recognition.

Possible future work directions of this paper involve creating new kinds of Romanian language models using our novel dataset, such as GPT-2 \cite{radford2019language} for tweet generation. In addition, we intend to add the results obtained by our RoBERTweet models in LiRO \cite{dumitrescu2021liro}, a Romanian benchmark for NLP models.

\section*{Acknowledgements}
This research has been funded by the University Politehnica of Bucharest through the PubArt program.
We gratefully acknowledge the support of the TensorFlow Research Cloud program\footnote{\url{https://sites.research.google/trc}} for generously providing us access to Cloud TPUs, which enabled us to carry out our extensive pre-training experiments. 

\bibliographystyle{splncs04}
\bibliography{bibliography}

\end{document}